\documentclass[sigconf]{aamas}

\usepackage{balance} 

\usepackage{textcomp}
\usepackage{graphicx}
\usepackage[version=4]{mhchem}
\usepackage{siunitx}
\usepackage{longtable,tabularx}
\usepackage{accents}
\usepackage{comment}
\usepackage{mathtools}
\usepackage{subfigure}
\usepackage{caption}
\usepackage{color}
\usepackage{booktabs}
\usepackage{tabu}
\usepackage{algorithm}
\usepackage{algpseudocode}
\usepackage{multirow}
\usepackage{svg}
\usepackage{epstopdf}
\usepackage{makecell}
\usepackage{siunitx} 
\usepackage{enumitem}

\usepackage{standalone}

\usepackage[normalem]{ulem} 

\setcopyright{ifaamas}
\acmConference[AAMAS '25]{Under review for Proc.\@ of the 24th International Conference
on Autonomous Agents and Multiagent Systems (AAMAS 2025)}{May 19 -- 23, 2025}
{Detroit, Michigan, USA}{A.~El~Fallah~Seghrouchni, Y.~Vorobeychik, S.~Das, A.~Nowe (eds.)}
\copyrightyear{2025}
\acmYear{2025}
\acmDOI{}
\acmPrice{}
\acmISBN{}

\acmSubmissionID{<<1204>>}

\title{Resolving Multiple-Dynamic Model Uncertainty in Hypothesis-Driven Belief-MDPs}

\author{Ofer Dagan}
\affiliation{
  \institution{University of Colorado, Boulder}
  \city{Boulder}
  \country{United States}}
\email{ofer.dagan@colorado.edu}

\author{Tyler Becker}
\affiliation{
  \institution{University of Colorado, Boulder}
  \city{Boulder}
  \country{United States}}
\email{tyler.becker-1@colorado.edu}

\author{Zachary N. Sunberg}
\affiliation{
  \institution{University of Colorado, Boulder}
  \city{Boulder}
  \country{United States}}
\email{zachary.sunberg@colorado.edu}

\begin{abstract}

When human operators of cyber-physical systems encounter surprising behavior, they often consider multiple hypotheses that might explain it.
In some cases, taking information-gathering actions such as additional measurements or control inputs given to the system can help resolve uncertainty and determine the most accurate hypothesis.
The task of optimizing these actions can be formulated as a belief-space Markov decision process that we call a \emph{hypothesis-driven belief MDP}.
Unfortunately, this problem suffers from the curse of history similar to a partially observable Markov decision process (POMDP).
To plan in continuous domains, an agent needs to reason over countlessly many possible action-observation histories, each resulting in a different belief over the unknown state.
The problem is exacerbated in the hypothesis-driven context
because each action-observation pair spawns a different belief for each hypothesis, leading to additional branching.  
This paper considers the case in which each hypothesis corresponds to a different dynamic model in an underlying POMDP.  
We present a new belief MDP formulation that: 
(i) enables reasoning over multiple hypotheses, 
(ii) balances the goals of determining the (most likely) correct hypothesis and performing well in the underlying POMDP, and
(iii) can be solved with sparse tree search.

\end{abstract}


\newcommand{\BibTeX}{\rm B\kern-.05em{\sc i\kern-.025em b}\kern-.08em\TeX}

\begin{document}


\pagestyle{fancy}
\fancyhead{}


\maketitle 

\section{INTRODUCTION}
\label{sec:intro}

In sequential decision-making problems modeled as a Partially Observable Markov Decision Processes (POMDPs), an autonomous agent must account for uncertainty.  
To reason about uncertainty, the agent maintains a belief, or a probability distribution over the unknown state of the system, by updating its prior belief using Bayes' rule with noisy transition and measurement models.
When planning in continuous domains, this results in a different belief for each possible action-observation pair, making the POMDP problem intractable, also known as the `curse of history'. 
Further growth in complexity occurs due to ambiguity, e.g., uncertainty in which object generated the measurement (i.e., the data association problem), or uncertainty in which dynamic model generated the state transition.
In these cases, for the same observation or action, multiple `branches' of the belief must be reasoned over, or estimated.

In the Bayesian estimation literature, multi-hypothesis algorithms are separated based on the origin of the hypothesis.
In the first set of problems branching of the belief space occurs in the prediction step, due to multiple-dynamic models. 
These are addressed by algorithms such as the multiple-model (MM) and the interactive-MM (IMM) algorithms \cite{bar-shalom_adaptive_2001}, \cite{ristic_beyond_2003}.
In the second set of problems, branching occurs in the measurement update step, due to ambiguity, where there is uncertainty in which object generated the measurement signal, also known as the data association problem.
Algorithms for this set of problems include multi-hypothesis tracking (MHT) \cite{blackman_design_1999}, and the probabilistic data association filter (PDAF) and joint probabilistic data association (JPDA) filters \cite{bar-shalom_probabilistic_2009}.
While these problems are similar in that they handle uncertainties and multiple potential scenarios, they are complimentary problems, and the solutions are not interchangeable. 
A similar distinction can be made in multi-hypothesis planning algorithms, sometimes referred to as `hybrid belief' planning, referring to the belief containing continuous and discrete variables, between methods that consider measurement ambiguity, and ones that consider multiple-dynamic models.

\textbf{Measurement ambiguity:} 
Planning with measurement ambiguity has received increased attention in the active SLAM (simultaneous localization and mapping) community.
A Monte-Carlo (MC)--based algorithm for hybrid belief POMDPs (HB-MCP) is presented by \citet{barenboim_monte_2023}.
It builds on the DA-BSP and D2A-BSP algorithms \cite{pathak_unified_2018}, \cite{shienman_d2a-bsp_2022}, and extends it to prune unlikely hypotheses and keeping the most promising ones.
The ARAS algorithm \cite{hsiao_aras_2020} uses factor graphs with multi-hypothesis factors \cite{hsiao_mh-isam2_2019} for ambiguity-aware active SLAM, but chooses only one hypothesis to plan the next path segment.\\
\textbf{Multiple-dynamic models:} 
Planning with hybrid dynamics is addressed by \citet{brunskill_continuous-state_2008} and \citet{jain_belief_2017}, where the idea is to approximate a non-linear dynamics model using a set of linear models. 
The approach of \citet{brunskill_continuous-state_2008} is limited to Gaussian sum filters for belief updates, and assumes discrete actions and observations spaces, thus does not consider the full breadth of the problem. 
Similarly, \citet{jain_belief_2017} represent the belief with a Gaussian mixture, but consider continuous action space with maximum likelihood observations.
In both types of multi-hypothesis planning, the key objective is based on the system's underlying state, such as the agent's position, the position of a tracked target, or a map of the environment.
This is done by taking the marginal of the system state from the joint belief, which can be thought of as taking a weighted average based on the likelihood of the possible hypotheses.
Thus, the task of determining which hypothesis is most probable is not considered, nor does it directly drive the optimization process.
Determining the correct hypothesis, however, can be crucial for explaining a surprising behavior or to understand certain outcomes.

For example, consider a human operator that detects an unexpected behavior while monitoring a cyber-physical system, e.g., a drone performing some inspection task while suddenly suffering from a possible rotor failure \cite{laouar_feasibility-guided_2024}, or a space-object tracking system for space domain awareness (SDA) that observes some anomaly in the orbit of an object \cite{jaunzemis_evidence_2019}, the operator might want to consider multiple hypotheses that can explain it. 
We define the task of optimizing actions that can help resolve uncertainty and determine the most probable hypothesis as \textit{hypothesis-driven planning}. 
In these problems, an autonomous agent must balance between actions that optimize the underlying planning problem, and determining the most likely hypothesis,  
in the drone example that means continuing with the inspection task vs. taking maneuver control actions to help determine whether a rotor failed. 

This paper focuses on the autonomous agent side of the team to formulate a new planning framework, the hypothesis-driven belief MDP.
In this formulation, the planning agent seeks a policy that balances two competing requirements: 1. solving an underlying POMDP problem, and 2. determining the correct hypothesis stemming from uncertainty over which transition model drives the state of interest.
The paper contributions are (i) presenting a new multiple-dynamic hypothesis belief MDP (MDH-BMDP) that enables adding multiple hypotheses for existing POMDP problems and reasoning over arbitrary belief shapes using existing sparse tree search solvers,  
(ii) suggesting a reward function that supports balancing the objectives of determining the correct hypothesis and performing well in the underlying POMDP, and 
(iii) demonstrating the applicability of the framework and performance of the new reward function in simulations.

\section{Background}
\label{sec:background}

\subsection{Belief-MDP}
One of the main building blocks for sequential decision-making under uncertainty is the Markov decision process (MDP). 
An MDP is defined by the tuple $(\mathcal{S}, \mathcal{A},\mathcal{T},\mathcal{R},\gamma)$, where $\mathcal{S}$, $\mathcal{A}$ are the sets of all possible states and actions, respectively.
$\mathcal{T}(s,a,s')=p(s'|s,a)$ is a stochastic state transition model, which defines the probability of transitioning to state $s'\in \mathcal{S}$ from state $s\in \mathcal{S}$ after taking action $a\in \mathcal{A}$. 
The reward function $\mathcal{R}(s,a)$ determines the immediate reward the agent receives when taking action $a$ at state $s$, and $\gamma\in[0,1)$ is a discount factor.

A solution for an MDP is an optimal
policy $\pi^*(s)$ that maps every state to an action $a^*$, and maximizes the expected cumulative reward. 
However, in most real-world problems, the state $s$ is not perfectly known, but can only be inferred by a sequence of noisy measurements $o$.
A partially observable MDP (POMDP) takes into account this uncertainty by adding two elements, $(\mathcal{O},Z)$ to the MDP tuple $(\mathcal{S}, \mathcal{A},\mathcal{T},\mathcal{R},\gamma)$. 
Where $\mathcal{O}$ is the set of all possible observations, and $Z(s',a,o)=p(o|s',a)$ is the stochastic observation function, defined as the conditional probability of seeing observation $o$ after taking action $a$ and reaching state $s'$.

As the true state $s$ of a POMDP is unknown, an agent maintains a \emph{belief} over states $s\in \mathcal{S}$, which summarizes the history $h_t$ of all actions taken and observations received up to and including time step $t$ and starting from a prior belief $b_0$,
\begin{eqnarray*}\label{eq:belief}
    b_t(s)=p(s_t=s|h_t), \ \ \textit{where} \ \ h_t=(b_0,a_0,o_1,a_1,...,o_t).
\end{eqnarray*}
A difficulty is that the reward function of a POMDP depends directly on the state of the system and, thus this reward cannot be used to account for state uncertainty or to motivate information-gathering actions \cite{araya_pomdp_2010}.
A solution to this problem is to formulate the problem as a belief-MDP and reason about belief-states instead of states. 
Note that any POMDP is a belief-MDP, with a set of belief states $b\in\mathcal{B}$,  a transition model $\mathcal{T}(b,a,b')$ \cite{kaelbling_planning_1998}, which accounts for observations using Bayes' rule,
\begin{eqnarray}\label{eq:transition}
    \mathcal{T}(b,a,b') = p(b'|b,a)=\int_{o\in\mathcal{O}} p(b'|a,b,o)\cdot p(o|a,b),
\end{eqnarray}
and a belief-dependent reward function \cite{araya_pomdp_2010}
\begin{eqnarray}\label{eq:rhoReward}
    \rho(b,a)=\mathbb{E}_{s\sim b}\big [ \mathcal{R}(s,a)\big ]+\varrho(b,a).
\end{eqnarray}
Here $\varrho(b,a)$ is a belief-dependent reward such as Shannon's entropy or Kullback-Leibler divergence, and the first term is the expected reward when the state $s$ is distributed according to $b$,  
\begin{eqnarray}{\label{eq:expctedR}}
    \mathbb{E}_{s\sim b}\big [ \mathcal{R}(s,a)\big ]=\int_{s\in \mathcal{S}}b(s)\mathcal{R}(s,a).
\end{eqnarray}

\subsection{MCTS}
\begin{figure*}[htb!]
    \centering
    \includegraphics[width=0.98\textwidth]{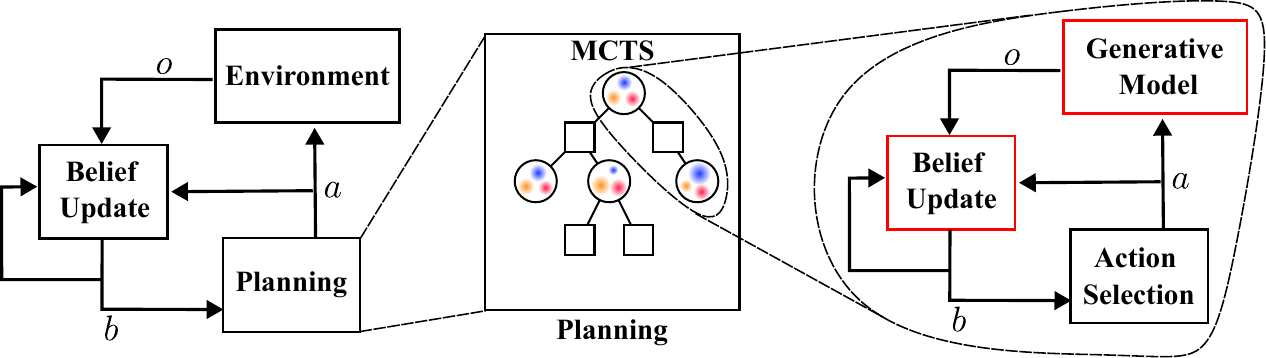}
    \caption{Plannig - Acting - Observing - Estimating loop diagram and how it translates to the MCTS algorithm.}
    \label{fig:MCTS_diagram}
    \Description{MCTS_diagram}
\end{figure*}

A sequential decision process (loop) is built out of four main steps: plan -- act -- observe -- estimate\footnote{There are different versions and terms relating to a decision process in the literature, here we choose terms that relate to Bayesian estimation and decision making.}.
The left diagram in Figure \ref{fig:MCTS_diagram} describes a cycle in an online planning loop, where an agent plans based on its prior belief $b$ to choose an action $a$, acting and interacting with the environment, and generating a measurement $o$. 
The last step is a Bayesian belief update,
\begin{eqnarray}\label{eq:belUpdate}
    b'(s') = p(s'|h_t)\propto p(o|a,s') \cdot \int_s b(s)\cdot p(s'|s,a),
\end{eqnarray}
where $s$ is the state at the current time step and $s'$ represents the state at the next time step.

Monte Carlo tree search (MCTS) is a widely used technique to solve various reasoning problems. 
In the context of Belief-MDP problems, at each planning step, MCTS builds a tree, consisting of belief (circle) and action (square) nodes, as shown in Figure \ref{fig:MCTS_diagram} (middle), up to a predefined horizon. 
The result of the simulation is the (current step) action that maximizes the action value function,
\begin{eqnarray}{\label{eq:Qvalue}}
    Q(s,a)=\mathcal{R}(s,a)+\gamma\sum_{s'}\mathcal{T}(s'|s,a)U(s'),
\end{eqnarray}
where $U(s)=\max_a Q(s,a)$ is the value function \cite{kochenderfer_algorithms_2022}. 
Since the MCTS algorithm is not the focus of this paper, we now discuss the elements of the algorithm that are relevant in the context of this paper and refer the interested reader to the survey by \citet{browne_survey_2012} for more details about the MCTS algorithm and to the documentation of the POMDPs.jl framework~\cite{egorov_pomdpsjl_2017} for implementation details.

Figure \ref{fig:MCTS_diagram}(right) shows a process diagram of one expansion of the tree, that is from a prior belief $b$ to a posterior belief $b'$.
Note that after taking an action $a$ from belief $b$, there are many possible posterior beliefs $b'$, each conditioned on a different possible observation $o$. 
The two MCTS elements pertinent to this paper are the generative model and the Bayesian belief update.
In the context of Markov decision processes, a generative model $G(\mathcal{P},s,a)$ is a function that simulates the environment, that is, given a problem tuple, $\mathcal{P}$, a state, and an action, it generates the next state $s'$, an observation $o$, and a reward $r$ as described in algorithm \ref{al:gen_model_S}.

Assuming a Bayesian approach, a belief update refers to the process in which a prior distribution is propagated according to the system dynamics and then updated according to a measurement (evidence), as described in Equation \ref{eq:belUpdate}.
While the key update steps are general, the implementation differs based on the type of underlying distribution, its representation, and the functions $\mathcal{T}$ and $Z$.
Further discussion regarding the type of updater suited for the multiple-dynamic hypothesis problem is given in section \ref{sec:updater}.

\begin{algorithm}[bth!]
\algrenewcommand\algorithmicprocedure{\textbf{algorithm}}
\small
\caption{\small Generative Model $G(\mathcal{P},s,a)$}
\label{al:gen_model_S}

\begin{algorithmic}[1]
\State \textbf{Input:} $s$, $a$,  $\mathcal{P} = \{ \mathcal{S}, \mathcal{A},\mathcal{T},\mathcal{O}, Z, \mathcal{R},\gamma\}$
\State $\mathcal{T} = \mathcal{P}.\mathcal{T}$, \ $Z = \mathcal{P}.Z$, \ $\mathcal{R} = \mathcal{P}.\mathcal{R}$
\State $s' \sim \mathcal{T}(s'|s,a)$
\State $o \sim Z(o | a, s')$
\State $r \leftarrow \mathcal{R}(s, a, s', o)$
\State \Return $(s', o, r)$
  
\end{algorithmic}

\end{algorithm}



    

\section{Problem Statement}
\label{sec:prob_statement}

Consider an autonomous agent, tasked with solving an underlying partially observable Markovian planning problem $\mathcal{P}$, that can be framed as a POMDP, and thus a belief-MDP.
Denote the underlying state space of the system as $\mathcal{S}_x$ and the corresponding belief space as $\mathcal{B}_x$.
Now assume that there is uncertainty regarding the transition model that derives 
the underlying state space (or a subset of it), that is, there are $n_{\mathcal{H}}$ possible transition models $\mathcal{T}_i$ (with no transitions between the models). 
We define the set of questions of whether model $i$ is correct or not as the set of hypotheses $\mathcal{H}$, where each hypothesis $\mathcal{H}_i$ corresponds to transition model $\mathcal{T}_i$.
Then the belief over all possible hypotheses is represented by $\mathcal{B}_{\mathcal{H}}$.

We define the new multiple-dynamics hypothesis (MDH) problem $\Bar{\mathcal{P}}$, as a planning problem over both the underlying belief state $\mathcal{B}_x$ and the hypothesis state $\mathcal{B}_{\mathcal{H}}$, with the MDH-belief-MDP (MDH-BMDP) defined by the tuple,
\begin{eqnarray}{\label{eq:bmdpTuple}}
    (\bar{\mathcal{B}}=\mathcal{B}_x\times\mathcal{B}_{\mathcal{H}}, \mathcal{A},\bar{\mathcal{T}}, \bar{\rho},\gamma),
\end{eqnarray}
where $\mathcal{A}$ and $\gamma$ are the same as in the underlying problem $\mathcal{P}$, 
$\bar{\mathcal{B}}$ is the augmented belief-state set and includes the underlying belief-state space and the belief over all possible hypotheses.
The transition model $\bar{\mathcal{T}}$ includes a set of transition models $\mathcal{T}_i$ for $i=1:n_{\mathcal{H}}$ for all dynamics hypotheses, and $\bar{\rho}$ is the new reward function, which depends on both the underlying problem reward, $\rho_{x}$ and the hypothesis-based reward $\rho_{\mathcal{H}}$,
\begin{eqnarray}{\label{eq:bmdpRho}}
    \bar{\rho} = \rho_{x}(b_x,a)+\rho_{\mathcal{H}}(b_{\mathcal{H}},a).
\end{eqnarray}
The observation space and function, $\mathcal{O}, Z$, are not explicitly defined in the Belief-MDP tuple, as the MDP formulation assumes perfect state knowledge. 
Instead, they are implicitly included within the transition model $\bar{\mathcal{T}}$. 

The planning problem $\Bar{\mathcal{P}}$ considered in this paper searches for an optimal policy $\pi^*$ that balances two potentially competing requirements -- deciding which hypothesis $\mathcal{H}_i$, corresponding to transition model $\mathcal{T}_i$, is (most likely) correct, while still solving the original underlying problem $\mathcal{P}$.
This is different from existing works such as \cite{jain_belief_2017}, \cite{brunskill_continuous-state_2008} which search for a policy assuming the state can transition according to different dynamics and do not explicitly try to reason which one is correct.




\section{Technical Approach}
\label{sec:tech_app}
Typically, Belief-MDP (and POMDP) problems reason about one transition model $\mathcal{T}$, and one observation function $Z$ associated with the problem definition tuple $\mathcal{P}$.
This section describes the MDH-BMDP framework to enable the use of existing algorithms to solve the hypothesis-driven problem where several transition models are considered. 
We first define how the MDH-BMDP tuple, defined in Equation (\ref{eq:bmdpTuple}), is constructed given an underlying POMDP, and a set of hypotheses.
Second, we describe some important implementation details and then provide further details on the definition of the transition model $\bar{\mathcal{T}}$ and the reward function $\bar{\rho}$.

\subsection{Constructing the MDH-BMDP  }

\begin{figure}[tb!]
    \centering
    \includegraphics[width=0.48\textwidth]{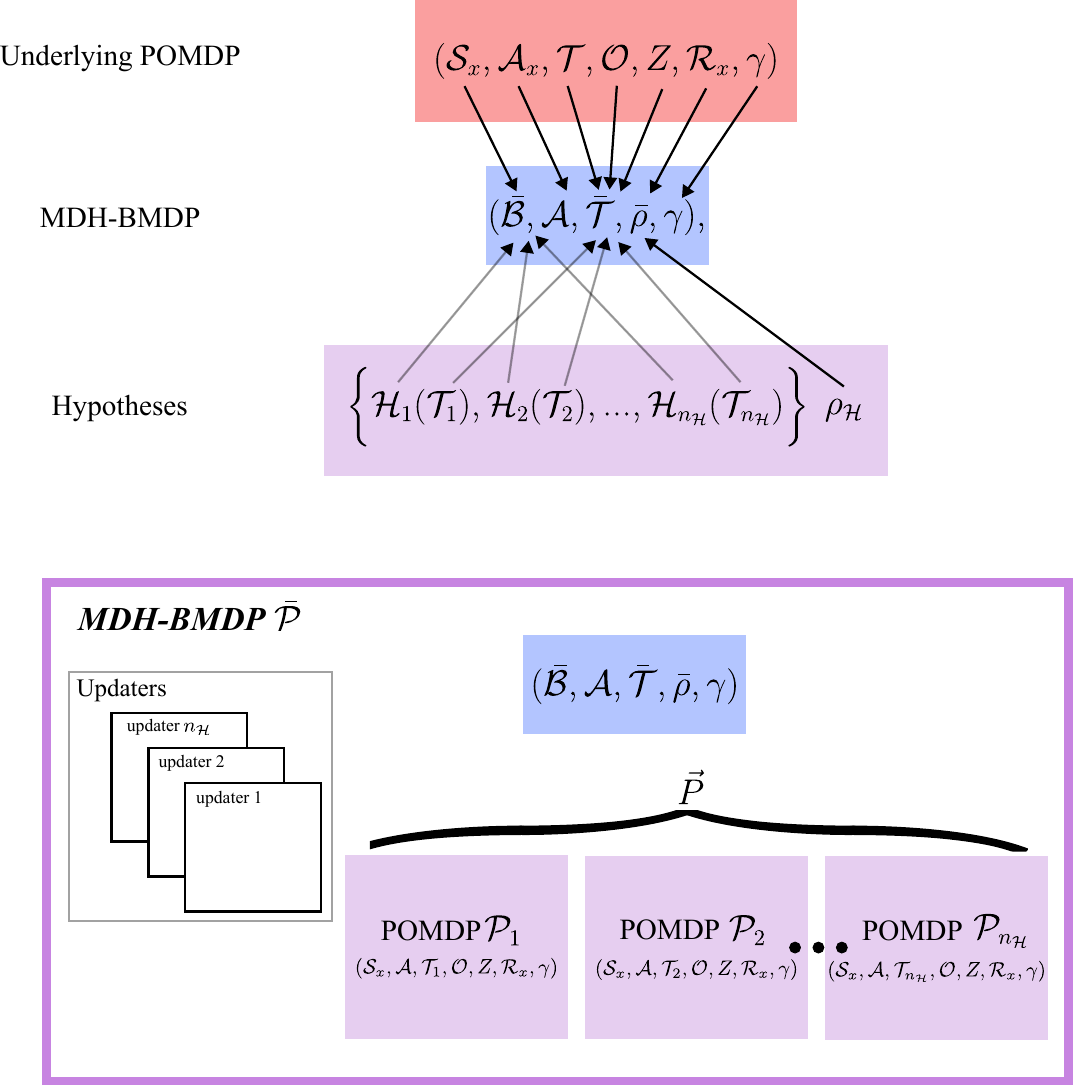}
    \caption{MDH-BMDP framework diagram }
    \label{fig:MDH_diagram}
    \Description{A diagram describing the structure of the MDH-BMDP framework. Shown are connections between the different components.}
\end{figure}

Consider an underlying POMDP problem $\mathcal{P}$, defined by $(\mathcal{S}_x, \mathcal{A},\mathcal{T}\\
,\mathcal{O}, Z, \mathcal{R}_x,\gamma)$, a set of $n_\mathcal{H}$ hypotheses that we want to reason about defined by their transition models,
\begin{eqnarray*}
    \bigl\{\mathcal{H}_1(\mathcal{T}_1), \mathcal{H}_2(\mathcal{T}_2),...,\mathcal{H}_{n_\mathcal{H}}(\mathcal{T}_{n_\mathcal{H}})\bigl\},
\end{eqnarray*}
and an hypotheses belief reward function $\rho_\mathcal{H}$.  
The elements of the MDH-BMDP definition $(\bar{\mathcal{B}}, \mathcal{A},\bar{\mathcal{T}}, \bar{\rho},\gamma)$ (Equation (\ref{eq:bmdpTuple})), and their dependencies  
on the underlying POMDP $\mathcal{P}$, the hypotheses set and the hypotheses belief reward function, are illustrated in Figure \ref{fig:MDH_diagram}(top) and defined as follows:
\begin{itemize}
    \item $\bar{\mathcal{B}}=\mathcal{B}_x\times\mathcal{B}_{\mathcal{H}}$ is the joint belief space, based on the underlying state space $\mathcal{S}_x$ and the hypotheses state space $\mathcal{S}_\mathcal{H}=1,2,...,n_\mathcal{H}$.
    As a result a belief state $\bar{b}\in \bar{\mathcal{B}}$ is given by,
\begin{eqnarray}{\label{eq:hybridBel}}
    \bar{b}(\bar{s}) = p(s_x, s_\mathcal{H}) = \underbrace{p(s_\mathcal{H})}_{{b}_\mathcal{H}}\cdot \underbrace{p(s_x|s_\mathcal{H})}_{{b}_x}.
\end{eqnarray}
Where ${b}_\mathcal{H}$ is a categorical distribution, and ${b}_x$ is defined as an array of conditional distributions, e.g. Gaussian, one for each hypothesis $\mathcal{H}_i$.
Thus the joint belief $\bar{b}$ is a mixture of distributions, weighted by the probability of each hypothesis $\mathcal{H}_i$ being correct.
    \item $\mathcal{A}$ is the action space, inherited from the underlying POMDP problem $\mathcal{P}$.
    \item $\bar{\mathcal{T}}$ is a generative transition model, defining how to generate a new joint belief $\bar{b}'$, from $\bar{b}'$, after taking an action $a$.
    More details about $\bar{\mathcal{T}}$ are given in Section \ref{sec:updater}, but notice that $\bar{\mathcal{T}}$ depends on the underlying transition model $\mathcal{T}$, observation space $\mathcal{O}$, and observation function $Z$, and the set of hypotheses transition models $\mathcal{T}_1,...\mathcal{T}_{n_\mathcal{H}}$
    \item $\bar{\rho}$ is the joint belief-dependent reward function defined in Equation (\ref{eq:bmdpRho}). It transitions the underlying state-dependent reward $R_x$ to a belief-dependent reward using Equation (\ref{eq:expctedR}) and adds the hypothesis reward $\rho_\mathcal{H}$. 
    Further discussion on the reward function is given in Section \ref{sec:reward}.
    \item $\gamma$ is the discount factor, inherited from the underlying POMDP problem $\mathcal{P}$.
\end{itemize}

\subsection{Implementation Details}
The key idea in the implementation of the MDH-BMDP framework is to define an array $\Vec{P}$ of  
$n_\mathcal{H}$ ``hypothesis conditioned''-POMDPs $\mathcal{P}_i$, 
\begin{eqnarray}
    (\mathcal{S}_x, \mathcal{A},\mathcal{T}_i,\mathcal{O}, Z, \mathcal{R}_x,\gamma),
\end{eqnarray}
where, $\mathcal{T}_i$ is based on the underlying POMDP problem $\mathcal{P}$, but with a dynamics function $f_i$ corresponding to hypothesis $\mathcal{H}_i$.
Note the subtle difference between the dynamics function $f_i$ and the transition model $\mathcal{T}$ -- while $f_i$ describes the equations of motion, $\mathcal{T}$ depends on $f_i$ and represents a distribution over possible states $s'_x$. 
The rest of the tuple elements are inherited from $\mathcal{P}$.
To be able to use the different POMDPs $\mathcal{P}_i$ with existing solvers within the POMDPs.jl ecosystem~\cite{egorov_pomdpsjl_2017}, during construction, the MDH-BMDP is augmented with the POMDP array $\Vec{P}$ and an array of belief updaters, one for each model $\mathcal{P}_i$.
We denote this new structure as $\bar{\mathcal{P}} = \{ (\bar{\mathcal{B}}, \mathcal{A},\bar{\mathcal{T}}, \bar{\rho},\gamma), \vec{P}, \text{Updaters}\}$.
The full structure of the MDH-BMDP is demonstrated in Figure \ref{fig:MDH_diagram}(bottom).
As will be discussed in the next section, this allows for the belief update of the multi-hypothesis belief $\bar{b}$.



\subsection{Generative Transition Model}
\label{sec:updater}
\begin{algorithm}[bt!]
\algrenewcommand\algorithmicprocedure{\textbf{algorithm}}
\small
\caption{\small Belief Generative Model}
\label{al:gen_model}

\begin{algorithmic}[1]
\State \textbf{Input:} $\bar{b}$, $a$, $\bar{\mathcal{P}} = \{ (\bar{\mathcal{B}}, \mathcal{A},\bar{\mathcal{T}}, \bar{\rho},\gamma), \vec{P}, \text{Updaters}\}$
\State $\bar{\rho} = \bar{\mathcal{P}}.\bar{\rho}$
\State $b_x = \bar{b}.b_x$, $b_\mathcal{H} = \bar{b}.b_\mathcal{H}$
\State $s_\mathcal{H} \sim b_\mathcal{H}$
\State $s_x \sim b_x[s_\mathcal{H}]$
\State $o\leftarrow G(\vec{P}[s_\mathcal{H}],s_x,a)$ \Comment{{\color{Gray} Algorithm \ref{al:gen_model_S} }} 
\State $\bar{b}'\leftarrow \textit{update}(\bar{\mathcal{P}}, \bar{b}, a, o) $ \Comment{{\color{Gray} Algorithm \ref{al:update} }}
\State $r \leftarrow \bar{\rho}(\bar{b}, a, \bar{b}', o)$
\State \Return $(b', o, r)$
  
\end{algorithmic}
\end{algorithm}

Planning in belief-MDPs requires reasoning belief-states instead of over states, thus a generative model as defined in algorithm \ref{al:gen_model_S} 
has to be modified to propagate the belief $b$ and not the state $s$, as described in algorithm \ref{al:gen_model}.
Propagating a prior belief $b$ to a posterior belief $b'$ in a Bayesian belief updater (estimator) includes two main steps: prediction and correction (Equation \ref{eq:belUpdate}).
In the prediction step, the belief is propagated according to the transition model $\mathcal{T}(s,a,s')=p(s'|s,a)$. 
In the correction step, also known as measurement update, Bayes' rule is used to fuse an observation according to the model's observation model $Z(s',a,o)=p(o|s',a)$.
However, since we are reasoning over a belief state, we don't have a unique state $s'$ to sample an observation, as in line 4, algorithm \ref{al:gen_model_S}. 
Instead, we need to first sample a state from the belief, by first sampling a hypothesis state $s_\mathcal{H}$ out of ${b}_\mathcal{H}$, and then an underlying state $s_x$ is sampled from the conditional distribution $p(s_x|s_\mathcal{H})$ (lines 4-5).
An observation is generated using Algorithm \ref{al:gen_model_S}, called on the POMDP $\mathcal{P}_i$, corresponding to the sampled hypothesis state $s_\mathcal{H}$.

We can see here that the structure of the MDH-BMDP allows for the use of the hypothesis-relevant transition and observation models within the generative model $G$.
Similar use is done within the update and reward calls on lines 7-8, respectively.
In the case of MDH-BMDP, the joint belief $\bar{b}$ is defined as in Equation (\ref{eq:hybridBel}), where ${b}_\mathcal{H}$ is a categorical distribution,
Thus $\bar{b}$ is a mixture of distributions. 
The belief updater is then of multiple-model (MM) type, with the key steps described in Algorithm \ref{al:update}. 
More details regarding different versions of MM estimators can be found in \cite{ristic_beyond_2003}, where for the examples presented in this paper we implemented MM particle filter (MMPF), and Gaussian mixture (GM) updaters.
Without loss of generality, we assume from now on that the conditional distributions ${b}_x$ are of the same type and are thus updated in the same manner (e.g., all using a Kalman filter or a particle filter), but since each updater $i$ in the array of updaters is connected to a specific POMDP $\mathcal{P}_i$, this is not a constraint.

\begin{algorithm}[bth!]
\algrenewcommand\algorithmicprocedure{\textbf{algorithm}}
\small
\caption{\small MM Belief Update}
\label{al:update}

\begin{algorithmic}[1]
\State \textbf{Input:} $\bar{\mathcal{P}}, \ \bar{b}, \ a,\ o$
\State $b_x = \bar{b}.b_x$, $b_\mathcal{H} = \bar{b}.b_\mathcal{H}$
\For {$i=1,\ldots, n_\mathcal{T}$}
\Statex  // Update conditional belief over the underlying state:
\State $\textit{updater} = \bar{\mathcal{P}}.\textit{Updaters}[i]$
\State ${b}_x'[i]\leftarrow\textit{predict}(\textit{updater},{b}_x[i]$, a)
\State $\bar{b}'.{b}_x'[i]\leftarrow\textit{correct}(\textit{updater},{b}_x'^-[i]$, a, o)
\Statex  // Correct belief over hypotheses:
\State ${b}_\mathcal{H}'[i] \leftarrow {b}_\mathcal{H}[i]\cdot \int_{x\sim b'_x[i]} p(o|\mathcal{H}_i, x)dx $
\EndFor
\State $\bar{b}'.{b}_\mathcal{H}'\leftarrow {b}_\mathcal{H}'/\sum_i {b}_\mathcal{H}'[i]  $ \Comment{{\color{Gray} renormalize}}

\State \Return $\bar{b}' $
  
\end{algorithmic}

\end{algorithm}

\subsection{Reward Function }
\label{sec:reward}
The last element in the definition of the MDH-BMDP $\bar{\mathcal{P}}$ is the reward function $\bar{\rho}$. 
The goal is to construct a reward function that will balance two requirements: 
(i) making a decision regarding the most probable hypothesis, and 
(ii) solving the original underlying problem $\mathcal{P}$ with minimal effect on the original policy.

Similar to \cite{fischer_information_2020}, we define the reward function as the sum of the expected state-action reward and a belief-dependent reward as in Equation \ref{eq:bmdpRho}, 
\begin{eqnarray}{\label{eq:bmdpRho2}}
    \bar{\rho}(\bar{b},a) = \int_{s_x\in \mathcal{S}_x}b_x(s_x)\mathcal{R}_x(s_x,a) ds_x +w\cdot \rho_{\mathcal{H}}(b_{\mathcal{H}},a),
\end{eqnarray}
where we used Equation (\ref{eq:expctedR}) for the underlying POMDP reward, and added the parameter $w$ to allow weighting to balance between the two requirements. 
Since the reward function, $\mathcal{R}_x(s_x,a)$ is defined by the POMDP $\mathcal{P}_i$, $w$ allows tuning for the requested behavior of the MDH-BMDP policy. 
This is especially useful since for different underlying reward functions $\mathcal{R}_x$ the reward values might be scaled significantly differently.

We noe focus our attention on the hypothesis-related reward function $\rho_{\mathcal{H}}(b_{\mathcal{H}},a)$.
In a realistic scenario, where the agent is expected to generate a plan to support deciding which hypothesis is most likely, there are two parameters that could be of interest.
First, the user might want to make the decision within a time limit $\tau$.
Second, it is unlikely that the probability of one hypothesis will be exactly $1$, so we want to set a decision-making threshold, such that the probability of hypothesis $\mathcal{H}_i$ being correct is at least $p=1-\epsilon$.

Common belief-dependent reward functions often depend on some measure of information \cite{araya_pomdp_2010}, \cite{fischer_information_2020}, e.g., negative Shannon's entropy,
\begin{eqnarray}{\label{eq:shannonEnt}}
    \rho_\mathcal{H}(b_\mathcal{H},a)=\sum_{s_\mathcal{H}\in\mathcal{S}_\mathcal{H}}b_\mathcal{H}(s_\mathcal{H})\log b_\mathcal{H}(s_\mathcal{H}).
\end{eqnarray}
However, while this type of reward function can incentivize information-gathering actions that reduce entropy, thus trying to `push' one of the hypotheses probabilities in the direction of 1, it does not explicitly set a time limit on making a decision.
We suggest a new simple sparse reward function that explicitly accounts for making a decision of which hypothesis is correct with probability $p>1-\epsilon$ within time $\tau$. 
\begin{eqnarray}{\label{eq:sparseR}}
    \rho_\mathcal{H}(b_\mathcal{H},a)=
    \begin{cases}
1.0 & \text{ if } \ max(b_\mathcal{H})\geq 1-\epsilon \ \& \ t\leq \tau \ \& \ \text{!resolved}\\
0.0 & \text{ otherwise },  
\end{cases}
\end{eqnarray}
where `!resolved' should read `not resolved', and indicates that a decision had not been made yet, to avoid collecting the reward more than once. 
As might happen in cases when the probability fluctuates above and below the threshold. 




\section{Experiments}
\label{sec:Exp}

To test the suggested MDH-BMDP framework, and evaluate the effect of the reward function on hypothesis-driven BMDP problems, we performed simulation experiments on two problems -- the Van Der Pol (VDP) track problem, and a space domain awareness problem.
For both problems, we define an underlying POMDP, and then augment it with three hypotheses based on variations in the system dynamics.

\subsection{Multiple-Dynamic Model VDP-Track}
This test problem is a modification of the Van Der Pol (VDP) tag problem \cite{sunberg_online_2018-1} to a tracking application with multiple-dynamic model hypothesis. 
The underlying, or base planning problem of the autonomous agent is to plan an observation schedule for a sensor located at the origin, with the goal of estimating the unknown 2D position of three moving objects. 
The initial position of the objects is within a square with 0.5 unit side length, and they move according to the 2D VDP oscillation differential equations,
\begin{eqnarray}\label{eq:vdp}
    \dot{x} = \mu\bigg ( x-\frac{x^3}{3}-y\bigg ), \ \ \dot{y}=\frac{1}{\mu}x,
\end{eqnarray}
with $\mu=0.6/2.0/1.4$ for objects $1/2/3$, respectively.

A zero mean Gaussian process noise, with $\sigma_x=\sigma_y=0.05$, is added to the object's position at the end of each Runge-Kutta integration step (4th order).
The observation space is continuous and consists of 8 noisy range beams ($\mathcal{O} \in \mathbb{R}^8$) with $\sigma=2$.
In each time step the agent has one accurate measurement ($\sigma = 0.5$) at its disposal, and it needs to decide to which object it directs the sensor ($\mathcal{A} = \{1,\ 2, \ 3\} \in \mathbb{R}^3$), where there is a different probability of detection for each object, with $p_{detect} = 0.95\/ / 0.8\/ / 0.65$.
The reward for taking an accurate measurement of the object $i\in \{1,\ 2, \ 3\}$ is the Euclidean distance $d$ from the object.
Now consider a scenario in which an operator suspects that one of the objects, e.g., object 3, dubbed the object of interest (OOI), is not moving according to the assumed nominal model ($\mu=1.4$), and wants to check two other hypotheses regarding the dynamic model, that is with $\mu=3.0$, and $\mu = 0.75$. 
The state space is then 9-dimensional ($\mathcal{S}\in\mathbb{R}^9$), with the 2D position of the objects and 3 hypothesis states.
When transitioned into an MDH-BMDP, the belief state of the underlying VDPTrack POMDP is modeled using a particle filter with 250 particles for each hypothesis.
The VDPTrack MDH-BMDP problem $\bar{\mathcal{P}}$ balances between two competing requirements, (i) the base-POMDP task of estimating the objects' position, and (ii) determining the most likely dynamic model of the OOI, out of 3 possible dynamic models with probability $p=0.8$ and within 30-time steps.


\subsection{Space Domain Awareness}
One of the core problems in space domain awareness (SDA) is catalog maintenance.
It refers to the problem of tasking a set of sensors to maintain custody over an existing `catalog' of space objects (SOs).
This multi-sensor tasking problem can be formulated as a POMDP and solved via MCTS \cite{fedeler_sensor_2022}. 
Consider a scenario when during routine sensor tasking operation for catalog maintenance an anomaly in the orbit of one of the SOs is detected.
As a result, the operator is interested in determining the origin of the anomaly, e.g., due to an engine misfiring, a deployment of solar panels, or possibly an intentional maneuver.  

To demonstrate how MDH-BMDP is used for hypothesis-driven planning, we consider the following scenario: there are 5 SOs in low Earth orbit (LEO), and an anomaly is detected on one of them, the OOI, which spawns three hypotheses regarding the dynamic model: (i) nominal dynamics (i.e. nothing happened), (ii)-(iii) the OOI deployed a small/large solar panel which results in small/high amount of drag addition, respectively.  
Given a nominal catalog maintenance sensor-tasking plan $\mathcal{P}$ for 3600 seconds, with a time step of 60 seconds, the MDH-BMDP problem $\bar{\mathcal{P}}$ seeks to determine the correct hypothesis with probability $p=0.8$ within 1600 seconds, with minimal change to the original plan.
In this problem, when transitioned into an MDH-BMDP, the belief over the OOI state is approximated as a Gaussian distribution and propagated using an unscented Kalman filter (UKF) \cite{julier_new_1995}, one for each hypothesis.


\subsection{Results}
\begin{figure}[tb!]
    \centering
     \includegraphics[width=0.98\linewidth]{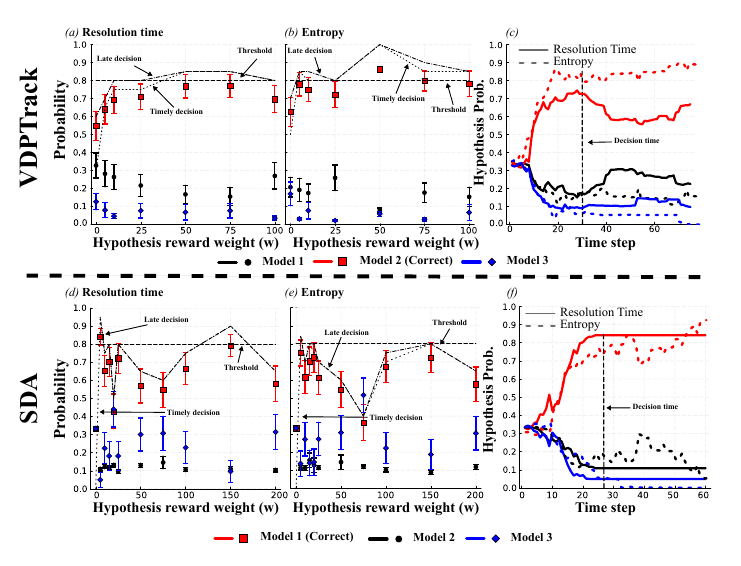}
    \caption{Hypothesis probability results for the VDPTrack (a)-(c) and the SDA (d)-(f) problems.}
    \label{fig:VDPRes}
    \Description{Results}
\end{figure} 

\begin{table*}[tbh!]
    \centering
    \caption{Result of chosen weights $w$ for VDPTrack and SDA problems, shown is the mean and standard error of the mean (SEM)}
	\label{tab:results}
    \begin{tabular}{c|c||c|c|c|c}\toprule
        Problem & $\rho_{\mathcal{H}}$ & Base reward &Steps to decide & \% Success rate (in time)  & \% Success rate (late)\\ \hline\hline
        \multirow{4}{*}{VDPTrack} & N/A ($w = 0$) & $52.9 \ (\pm0.84)$ & $30.5 \ (\pm2.94)$ & $34$ & $62$\\
         & Resolution Time $(w=50)$  & $\mathbf{51.3 \ (\pm 0.83)} $ & $12.2 \ (\pm0.85)$ & $84$ & $84$\\
         & Resolution Time $(w=75)$ & $50.1 \ (\pm0.74)$ & $\mathbf{12.1 \ (\pm0.81)}$ & $\mathbf{86}$ & $\mathbf{88}$ \\
         & Entropy $(w=50)$ & $46 \ (\pm0.66)$ & $14.2 \ (\pm1.11)$ & ${76}$ &  ${80}$\\ \hline\hline
        \multirow{3}{*}{SDA} & N/A ($w = 0$) & $0$  & N/A & $0$ & $0$ \\
         & Resolution Time $(w=150)$  & $\mathbf{-8.2 \ (\pm0.27)}$ & $14.3 \ (\pm0.93)$ & $\mathbf{90}$ & $\mathbf{90}$\\
         & Entropy $(w=150)$  & $-13.6 \ (\pm0.23)$  & $\mathbf{13.2\ (\pm1.18)}$  & $80$ & $80$ \\
    \end{tabular}
    
\end{table*}

In hypothesis-driven planning, the planning agent optimizes for actions that both reduce uncertainty and still perform well in the underlying problem. 
The simulation analysis then focuses on parameters relating to decision quality -- was the decision correct and was it made in time, and how much was the performance of the underlying problem affected -- as measured by the state-action dependent cumulative discounted reward (`base reward').

Figure \ref{fig:VDPRes} presents 20 Monte-Carlo (MC) simulation results of the multi-hypothesis VDPTrack and SDA problems, respectively.
The upper row of the figures shows a discrete search for the weighting factor $w$ of the hypothesis reward functions -- resolution time (Equation \ref{eq:sparseR}) and negative entropy (Equation \ref{eq:shannonEnt}).
For each weight $w$, subfigures (a)-(b) and (d)-(e) show (i) the discrete probabilities of the 3 different dynamic models, denoted by markers with error bars, at the time a decision was made, 
(ii) the threshold for decision making, $0.8$, in a dashed line, and (iii) success rate for making the right decision for a decision made in time (dots), and a decision made by the end of the simulation, including after the deadline, which we call `late' (dash-dot).
We can see then for the VDPTrack problem, the best results with respect to hypothesis decision are achieved with $w=50$ and $w = 75$ for the resolution time reward and $w=50$ for the entropy reward.
For the SDA problem, $w=150$ results in the best hypothesis decision results for both reward functions. 
Subfigures \ref{fig:VDPRes}(c) and (f) show the mean probability against time for $w=50$ and $w=150$, respectively, for both reward functions.
For the SDA problem, after a decision is made the resolution time reward function does not motivate the agent to take observation actions to the OOI, the probability stays constant. 
On the other hand, we can see that the entropy-based reward still causes fluctuations in the hypothesis probability. 
For the VDPTrack problem, since taking observations actions to the OOI might still be advantageous per the base reward function, we see that it might cause a decrease in the hypothesis probability, depending on the actual observations made.

To decide on a reward function and its best weight $w$, we consider how taking hypothesis-driven actions affects the underlying POMDP performance, we examine the base reward in Table \ref{tab:results}.
The baseline for comparison is a simulation with $w=0$, that is actions are taken based only on the base reward $\rho_x$.
For VDPTrack, we performed 50 MC simulations with the best-performing weights $w=[50, 75]$ (for resolution time) and $w=50$ (for entropy). 
From the table, we see that for both problems the resolution time reward outperforms the entropy-based reward since it leads to (i) a higher success rate for deciding on the correct hypothesis in time and (ii) a higher base reward.
The reward weight for the VDPTrack problem is chosen to be $w=50$, as it accumulates a higher base reward, thus having a smaller effect on the underlying problem, while still meeting the decision threshold and making a decision on time.  

For the SDA problem, the difference in the base reward can be explained by the number of unnecessary plan changes made when planning with the entropy reward.
In that case, there are $36.75$ plan changes on average versus only $14.25$ with the resolution time reward, where a decision is made after $13.2$ and $14.3$ steps on average, respectively. 
This means that the planner continues to choose actions to decrease hypothesis uncertainty, despite already making a decision, since the entropy reward has no mechanism to `inform' the planner that a decision was made. 
Note that this will happen naturally only in the unlikely event of the entropy being $0$, i.e., only when one hypothesis has probability $p=1.0$.

\begin{figure}[tb!]
    \centering
     \includegraphics[width=0.98\linewidth]{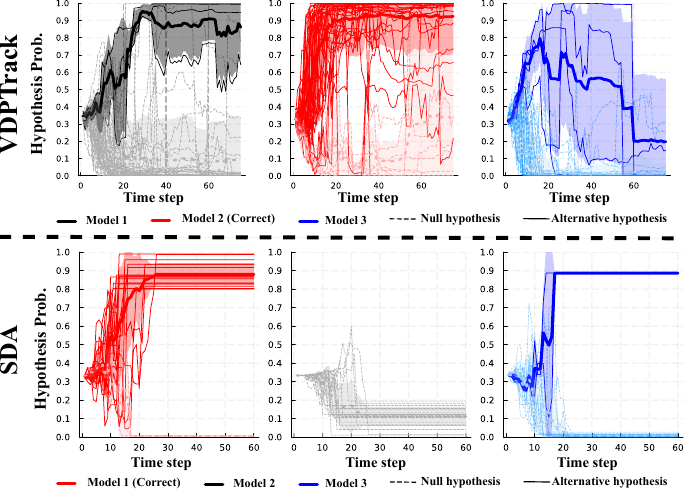}
    \caption{Probability traces for the VDPTrack (top row), and SDA (bottom row) problems.}
    \label{fig:SDARes}
    \Description{Results}
\end{figure}

For each dynamic model, there are two possible hypotheses, one is that the model does not describe the true dynamics of the OOI (the `Null' hypothesis), and the second is that the model describes the true dynamics (the `Alternative' hypothesis).
Figure \ref{fig:SDARes} presents probability traces of the two modes of each hypothesis - Null (dashed lines) and Alternative (full lines).
Results correspond to Table \ref{tab:results} with the resolution time reward function, with $w=50$ and $w=150$ for the VDPTrack and SDA problems, respectively.
As can be seen, for both problems, the algorithm decides on the right hypothesis, with the majority of the runs making the right decision -- the alternative hypothesis for the true model (in red), and the null hypothesis for the other models (black and blue). 

\section{Summary}
\label{sec:conclusion}

This paper defines the \textit{hypothesis-driven} POMDP problem as the set of multi-hypothesis POMDP problems where explicitly determining the most accurate hypothesis is one of the optimization objectives.
These problems naturally arise in human-robot collaboration, when the human detects a surprising behavior or unexpected outcome and wants to explore it.
In that case, reasoning over different possible models, that might justify the observed behavior and provide an alternative explanation. 
We formulate the problem as belief-MDP, dubbed MDH-BMDP, and structure it such that a given POMDP problem can be augmented with different hypotheses that relate to one or more of the problem's models, and then solved using existing sparse tree search algorithms.
While this paper focused on problems where hypotheses are instantiated due to multiple-dynamic models, the framework's architecture enables the exploration of other types of hypotheses.

\balance
To motivate actions that can help resolve uncertainty and determine the most probable hypothesis while still performing well in the underlying POMDP problem, we suggest a new reward function, explicitly rewarding in-time decisions.
Simulation results demonstrate the advantage of the new reward function over entropy-based reward, balancing between timely hypothesis decisions and the underlying problem objectives. 
Future work will look further into the effect of the hypothesis reward weight $w$ on the quality of the solution, aiming to develop intuition into the connection between the base reward and the hypothesis belief reward.

\begin{acks}
This work is supported by the Air Force Office of Scientific Research
(AFOSR), Grant number FA9550-23-1-0726.

\end{acks}



\bibliographystyle{ACM-Reference-Format} 
\bibliography{references}


\end{document}